\documentclass[runningheads]{llncs}

\usepackage{latexsym}
\usepackage[T1]{fontenc}
\usepackage[utf8]{inputenc}
\usepackage{microtype}
\usepackage{inconsolata}
\usepackage{multirow}
\usepackage{longtable}
\usepackage{graphicx}
\usepackage{CJKutf8}
\usepackage{tabularray}
\usepackage{float}
\usepackage{colortbl}

\begin{document}

\title{Gendec: A Machine Learning-based Framework for Gender Detection from Japanese Names}
\titlerunning{Gendec: Gender Dection from Japanese Names with Machine Learning}

\author{Duong Tien Pham\inst{1,2}\and
Luan Thanh Nguyen\inst{1,2,}\thanks{Corresponding author}
\\\textit{20521222@gm.uit.edu.vn, luannt@uit.edu.vn}}

\authorrunning{Pham et al.}

\institute{Faculty of Information Science and Engineering,\\ University of Information Technology, Ho Chi Minh City, Vietnam\and
Vietnam National University Ho Chi Minh City, Vietnam}

\maketitle

\begin{abstract}

Every human has their own name, a fundamental aspect of their identity and cultural heritage. The name often conveys a wealth of information, including details about an individual's background, ethnicity, and, especially, their gender. By detecting gender through the analysis of names, researchers can unlock valuable insights into linguistic patterns and cultural norms, which can be applied to practical applications. Hence, this work presents a novel dataset for Japanese name gender detection comprising 64,139 full names in romaji, hiragana, and kanji forms, along with their biological genders. Moreover, we propose \textbf{Gendec}, a framework for gender detection from Japanese names that leverages diverse approaches, including traditional machine learning techniques or cutting-edge transfer learning models, to predict the gender associated with Japanese names accurately. Through a thorough investigation, the proposed framework is expected to be effective and serve potential applications in various domains. 

\end{abstract}

\section{Introduction}

Gender detection based on human names task is known as a helpful tool for a broad range of fields, including sociological research, marketing, or personalization in technology applications \cite{blevins2015jane,roy2021demographical,wais2016gender}. While names appear straightforward, they include a wealth of cultural and historical complexities relating to gender identification. We can acquire significant insights into different areas of society and consumer behavior by recognizing biological genders through their human names. Thanks to the developments in natural language processing techniques, the gender recognition problem has developed over the years.

Current gender detection systems have utilized machine learning approaches to identify human gender based on their names automatically. Because of variances in linguistic properties of names in each language, various approaches can be examined to conduct the task of gender detection. To et al. \cite{to2020gender} published a Vietnamese dataset along with machine learning-based approaches toward gender prediction tasks for Vietnamese names. For Chinese human names, Jia et al. \cite{jia2019gender} conducted research to address the logosyllabic characteristic of Chinese characters, which also affects the probability of predicting gender. Furthermore, another study by Panchenko et al. \cite{panchenko2014detecting} proposed an efficient method for the Russian language. However, complex language systems with diverse alphabets, such as Japanese, still need datasets and experiments on this task.

In this paper, we introduce a Japanese names dataset with annotated gender labels serving the task of gender detection for the Japanese language. From the built dataset, we propose Gendec, a machine learning-based framework for gender detection from Japanese names, which aims to automatically predict the human gender of a given input name in the Japanese language. Moreover, a deeper investigation of Japanese names is also analyzed to address the Japanese language characteristics in terms of gender detection based on names.

The structure of this paper is as follows: First, Section 2 describes related works to our study, and the premises to conduct this research. Section 3 introduced the dataset proposed for the task of gender prediction with data analyses. Section 4 focuses on methodologies and introduces Gendec, a machine learning-based framework for gender detection based on Japanese names. Section 5 shows experiments in this research with experimental results and discussions. Section 6 concludes our work and draws future directions. 
\section{Related Work}

Due to linguistic variations, names can exhibit diverse traits, leading to numerous processing challenges. The way to sign a name to a person depends on their country culture and language characteristics. For people in the USA, naming a baby is relatively flexible compared to other countries, and that name can be inspired by various sources, including family names, cultural preferences, or popular trends \cite{larson2011naming,sigurd2008creativity}. In Russia, people have three names, including a given name, a patronymic, and a surname. The patronymic is formed from the given name of their father and is used to indicate the parentage of a person \cite{aksholakova2014proper,schochenmaier2021multicultural}. Besides, in Japan, naming a baby is rooted in tradition and cultural significance with the inherited family name. Moreover, the given name often carries meaning or reflects the family's wishes and can have various kanji characters, each with its meaning \cite{mori2020child}. This is why processing Japanese names is a challenging task.

Recently, machine learning and its models have been applied to several tasks about processing human names, especially gender detection based on names, to tackle limitations and improve detection performance. Hu et al. \cite{hu2021s} proposed a machine-learning approach to the task of English names by considering character-based machine-learning models as well as utilizing both first and last names as input. Another work from Ritesh et al. \cite{RITESH2018614} about word representation used deep learning for gender classification on names in various languages such as India, Western countries, Sri Lanka, and Japan. Their results showed the effectiveness of the word embeddings approach outperforming one-hot representation. Nastase et al. \cite{nastase2009s} presented an investigation of name-gender relations in German and Romanian. They proved the hypothesis with strong support by the high accuracy results from experiments based on the form of the words in names. However, for Japanese names, because of the existence of several ways to express a single romaji name in various kanji forms, there are still difficulties in processing their names \cite{ogihara2021know}. Hence, in this study, we aim to propose research about exploiting characteristics of Japanese names for gender detection. 
\section{Dataset}

Firstly, we build a specified dataset serving the task of gender detection with Japanese names based on the Japanese personal name dataset\footnote{https://github.com/shuheilocale/japanese-personal-name-dataset}, which only has separate first or last names. The final built dataset includes full Japanese names corresponding to their biological genders by romaji, hiragana, and kanji forms.

\subsection{Dataset Creation}
We extract individual kanji values from each data row in the first name set from the raw dataset, ensuring each row contains a unique set of kanji and corresponding romaji and hiragana. Next, these first-name data are incorporated with data in the last-name set for data augmentation. Note that we join first names and last names by the same gender correspondingly. The final dataset comprises 64,139 Japanese full-name samples in romaji, hiragana, and kanji forms, along with their corresponding genders. We divide the dataset with the proportions of 70, 20, and 10\% for training, validation, and test sets, respectively. The distribution of labels is relatively balanced, with approximately 49.84\% for male and 50.16\% for female samples. Table \ref{tab:dataset_sample} shows samples of the created dataset with full Japanese names and genders. The dataset we built for experiments in this paper can be found at our HuggingFace repository\footnote{https://huggingface.co/datasets/tarudesu/gendec-dataset}.

\begin{CJK}{UTF8}{min}
    \begin{table}[H]
    \centering
    \caption{Samples in the dataset with Japanese full names in romaji, kanji, and hiragana forms with their biological genders.}
    \label{tab:dataset_sample}
    \begin{tabular}{lllll}
    \hline
    \textbf{No.} & \textbf{Romaji Name} & \textbf{Kanji Name} & \textbf{Higarana Name} & \textbf{Gender} \\ \hline
    1 & Tamai Kazuyoshi & 玉井和善 & たまいかずよし & Male \\
    2 & Iwama Satoko & 岩間智子 & いわまさとこ & Female \\
    \hline
    3 & Shiraki \textcolor{red}{Yuka} & 白木\textcolor{green}{由花} & しらき\textcolor{red}{ゆか} & Female \\
    4 & Ikeno \textcolor{red}{Yuka} & 池野\textcolor{blue}{悠果} & いけの\textcolor{red}{ゆか} & Female \\
    5 & Sata \textcolor{cyan}{Kunishige} & 佐田\textcolor{yellow}{国重} & さた\textcolor{cyan}{くにしげ} & Male \\
    6 & Iso \textcolor{cyan}{Kunishige} & 磯\textcolor{magenta}{邦重} & いそ\textcolor{cyan}{くにしげ} & Male \\ \hline
    \end{tabular}
    \end{table}
\end{CJK}

The samples (3)-(4) and (5)-(6) in the above table show that the kanji form can be different despite having a similar romaji name. It indicates the diversity of homonymous expressions that a romaji name can have, which is one of the interesting characteristics of Japanese names that need to be addressed.

\subsection{Dataset Analysis}

\begin{figure}[H]
  \begin{minipage}{0.45\linewidth}
    \centering
    \includegraphics[width=1.0\linewidth]{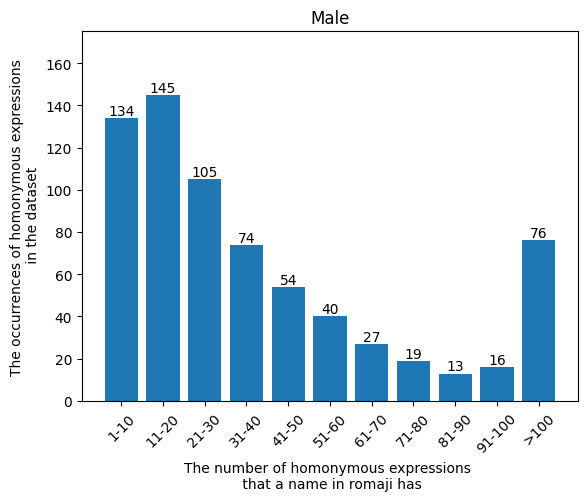}
    \label{fig:data_analysis_male}
  \end{minipage}
  \hspace{0.03\linewidth}
  \begin{minipage}{0.45\linewidth}
    \centering
    \includegraphics[width=1.0\linewidth]{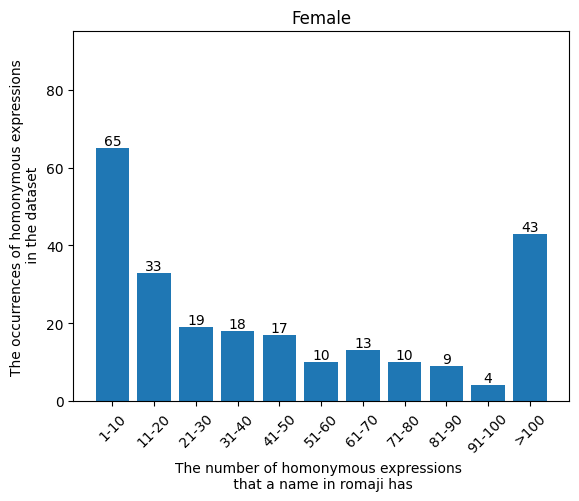}
    \label{fig:data_analysis_female}
  \end{minipage}
\caption{The analysis of the occurrences of homonymous expressions of a romaji word of male and female names, respectively.}
\label{fig:data_analysis}
\end{figure}

To analyze the dataset, we first conduct statistics of homonymous expressions that a romaji name can have for male and female names, respectively. Figure \ref{fig:data_analysis} illustrates that almost all names in the dataset have fewer homonymous expressions, around less than 20. However, there is still a high rate for the case of more than 100 homonymous expressions. Obviously, the diversity of homonymous expressions for one romaji name leads to homonym challenges in processing Japanese names.

\begin{CJK}{UTF8}{min}

\begin{figure}[H]
  \centering
  \begin{minipage}{0.45\linewidth}
    \centering
    \includegraphics[width=0.78\linewidth]{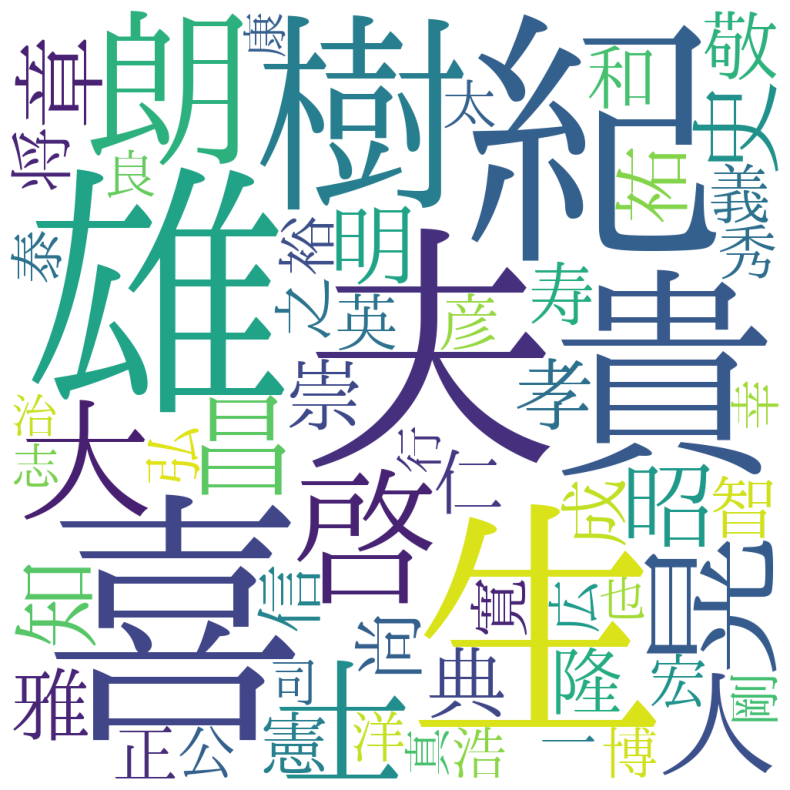}
    \caption{Word cloud of male names.}
    \label{fig:male_wc}
  \end{minipage}
  \hspace{0.0\linewidth} 
  \begin{minipage}{0.45\linewidth}
    \centering
    \includegraphics[width=0.78\linewidth]{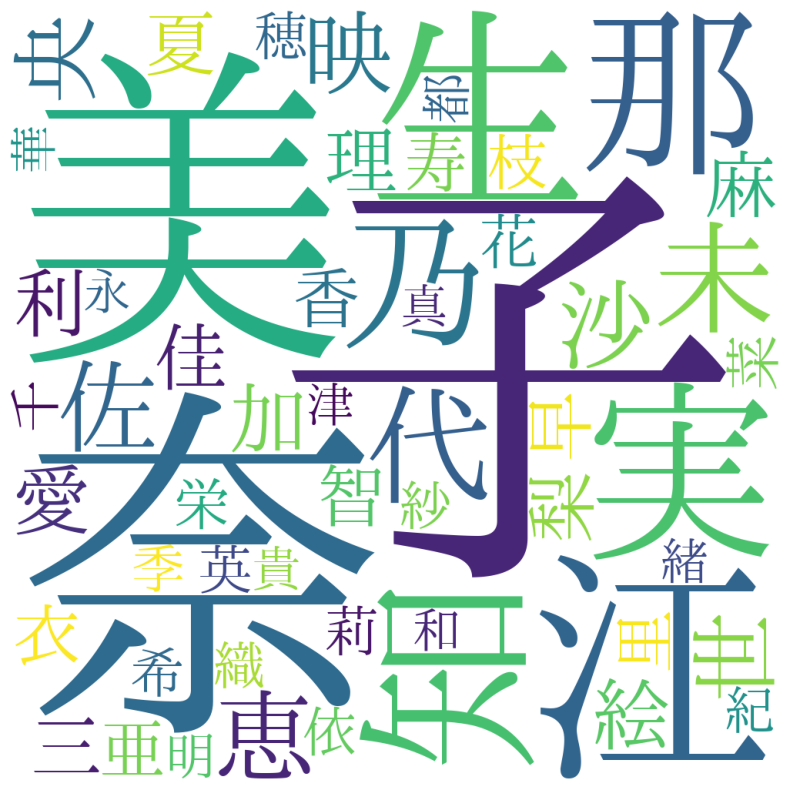}
    \caption{Word cloud of female names.}
    \label{fig:female_wc}
  \end{minipage}
\label{fig:word_cloud}
\end{figure}

Moreover, Figure \ref{fig:male_wc} and \ref{fig:female_wc} show word clouds of male and female Japanese first names in the dataset. 

We can see that the kanji characters 大 (big), 雄 (man), and 紀 (discipline) are the most used characters for naming men, and 子 (child), 美 (beauty), and 奈 (endurance) for women. It denotes the masculinity or femininity of a first name, perhaps aiding in gender recognition based on the human name. As indicated in Section 2, Japanese people's last names are named by inheriting their parents' and ancestors' names, which means that males and females might have similar last names, showing that the last name is not used to distinguish the gender of Japanese people.

\end{CJK}

\section{Methodologies}

\begin{figure}[h]
 \centering
 \includegraphics[width=1.\linewidth]{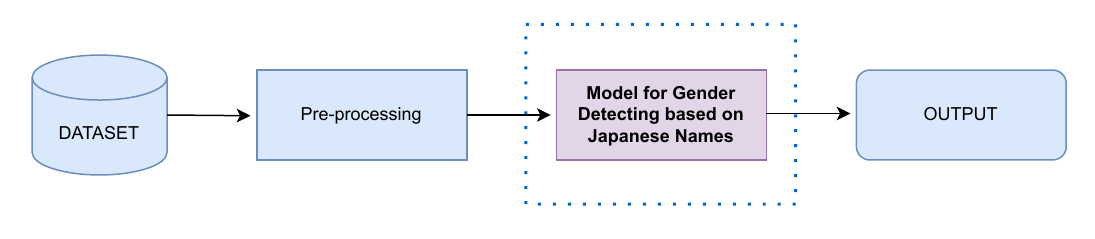}
 \caption{The overview of the proposed Gendec framework for the task of gender detection based on Japanese names.}
 \label{fig:overview}
\end{figure}

This study is followed by experiments conducted for the task of gender detection based on Japanese names with romaji- and kanji-form names in the built dataset. Figure \ref{fig:overview} below demonstrates the overview of our proposed Gendec framework.

\subsection{Pre-Processing Data}

Because of the simplicity of names, we only lowercase romaji-form names before feeding into the model. In addition, we deploy different ways of encoding data for traditional machine learning approaches, including TF-IDF and Count Vector, which effectively capture word frequencies and relationships within the data for improving the model's accuracy and performance. On the other hand, because of grasping linguistic semantics thanks to already pre-trained with tons of words, we only feed the raw lowercase romaji names into the model as the input data for the transfer learning approach.

\subsection{Gendec: Gender Detecting based on Japanese Names}
In this study, we aim to propose Gendec, a system for detecting gender based on Japanese names. The system takes a romaji-form name as input and outputs its predicted gender, including male or female, by utilizing the performance of text classification models. There are several options for the model in this research, comprising various approaches of traditional machine learning as well as transfer learning. 

\subsubsection{Traditional Machine Learning Approaches}

The task of gender detection is initially a binary classification task. Hence, we conduct the first experiments with various traditional machine learning methods to evaluate the dataset. 

\textbf{Support Vector Machine (SVM):} is an efficient machine learning method that can perform classification and regression tasks \cite{tong2001support}. It determines the optimum hyperplane for data separation, resulting in great performance and generalization. It is adaptive to both linear and non-linear data distributions, making it an excellent choice for a wide range of applications. 

\textbf{Naive Bayes:} is a probabilistic technique in machine learning that is commonly used for classification problems \cite{kim2006some}. It is based on Bayes' theorem and assumes feature independence, hence the "naive" designation. Despite its basic premise, Naive Bayes frequently outperforms other methods in practice, notably in text categorization. It computes the likelihood of an instance belonging to a given class based on the conditional probabilities of its characteristics, making it both computationally efficient and highly interpretable.

\textbf{Decision Tree:} is a well-known machine learning method for classification applications \cite{charbuty2021classification}. It iteratively splits the dataset into subsets depending on the most significant features, intending to create a decision-making tree structure. Each core node represents a feature-based judgment, whereas the leaf nodes include the final predictions or outcomes. Decision trees are highly interpretable and give information about the relevance of features.

\textbf{Random Forest:} is a strong ensemble machine learning technique that makes use of Decision Trees to boost prediction accuracy and durability \cite{xu2012improved}. It works by training several Decision Trees on distinct subsets of the data and using randomized feature selection. The algorithm then aggregates these trees' predictions by voting (for classification) or averaging (for regression) to create more accurate and stable predictions.

\textbf{Logistic Regression:} is a basic statistical and machine-learning approach for binary and multi-class classification applications \cite{shah2020comparative}. Unlike linear regression, it uses the logistic function to represent the likelihood of an instance belonging to a certain class, yielding values ranging from 0 to 1. Logistic Regression computes a decision boundary that divides the classes and estimates coefficients for input characteristics.

\subsubsection{Transfer Learning Approaches}
Besides evaluating the task with traditional machine learning models, we then implement transfer learning models to robust the performance of the proposed system. In this research, we choose multilingual transfer learning models, including mBERT, DistilmBERT, and XLM-R, for conducting experiments on this task of gender detection based on name.

\textbf{Bidirectional Encoder Representations from Transformers (BERT):} is a ground-breaking model for natural language comprehension and processing \cite{devlin2018bert}. BERT, which was introduced by Devlin et al. in 2018, has revolutionized several NLP activities. BERT employs a transformer architecture and has been pre-trained on enormous amounts of text data to learn contextual embeddings for words, allowing it to grasp language and meaning subtleties. BERT's bidirectional approach enables it to examine the complete context of a word inside a phrase, making it ideal for jobs requiring a grasp of word connections. BERT is regularly fine-tuned on specific tasks by researchers to attain cutting-edge performance. In experiments, we use the multilingual version of BERT called mBERT\footnote{https://huggingface.co/bert-base-multilingual-cased}.

\textbf{Distilled Bidirectional Encoder Representations from Transformers (DistilBERT):} is a lightweight and compressed form of the BERT model meant to make large-scale pre-trained models more accessible and computationally efficient \cite{sanh2019distilbert}. DistilBERT utilizes a distilled training strategy that mimics the behavior of the bigger BERT model while utilizing fewer parameters. As a result, the model is smaller and quicker, using less memory and computational resources, making it more suitable for diverse natural language comprehension and processing jobs. DistilmBERT is the multilingual version that is used in this study\footnote{https://huggingface.co/distilbert-base-multilingual-cased}.

\textbf{Cross-lingual Language Model - RoBERTa (XLM-R):} is a multilingual natural language processing model developed by Conneau et al. \cite{conneau2019unsupervised}. It was pre-trained on a wide, multilingual corpus, allowing it to comprehend and create text in various languages. XLM-R is a flexible tool for a wide range of NLP tasks, making it ideal for cross-lingual and multilingual applications and languages with limited training data or resources. We use the base version of XLM-R for the experiment\footnote{https://huggingface.co/xlm-roberta-base}.
\section{Experiments}

\subsection{Model Settings}
Our experiments utilized a single A100 GPU in the Google Colab environment\footnote{https://colab.research.google.com/} for all tasks. For fine-tuning transfer learning models, we set a value of 2e-5 for the learning rate, a total of 2 training epochs, and a batch size of 32. 

\subsection{Experimental Results}

After training these models, we achieve experimental results of the proposed framework Gendec, which uses various approaches, such as traditional machine learning or transfer learning models as classifiers, by macro F1 score on the test set. Table \ref{tab:experimental_results} demonstrates the results on the task of gender detection based on Japanese names in romaji form. Note that we train these models in two kinds of input data: the original romaji in the dataset and the converted romaji from the kanji name. Furthermore, we use names in available romaji from the test set to evaluate the performance of models.

\begin{table}[H]
\centering
\caption{The experimental results of various approaches on the task.}
\label{tab:experimental_results}
\resizebox{\textwidth}{!}{%
\begin{tabular}{llcccccc}
\hline
\multicolumn{2}{c}{\multirow{2}{*}{\textbf{Model}}} &
  \multicolumn{3}{c}{\textbf{Converted}} &
  \multicolumn{3}{c}{\textbf{Original}} \\ \cline{3-8} 
\multicolumn{2}{c}{} &
  \textbf{Female} &
  \textbf{Male} &
  \textbf{Average} &
  \textbf{Female} &
  \textbf{Male} &
  \textbf{Average} \\ \hline
\multirow{5}{*}{\begin{tabular}[c]{@{}l@{}}Traditional\\Machine Learning\\with TF-IDF\end{tabular}} &
  Decision Tree &
  86.96 &
  88.27 &
  87.62 &
  87.00 &
  88.26 &
  87.63 \\
                                   & Logistic Regression & 86.08          & 87.26          & 86.67          & 99.53          & 99.53          & 99.53          \\
                                   & Naive Bayes         & 79.20          & 64.51          & 71.86          & 97.80          & 97.69          & 97.75          \\
                                   & Random Forest       & \textbf{87.11} & \textbf{88.50} & \textbf{87.81} & \textbf{97.67} & \textbf{99.66} & \textbf{99.66} \\
                                   & SVM                 & 86.59          & 88.34          & 87.47          & 99.59          & 99.59          & 99.59          \\ \hline
\multirow{5}{*}{\begin{tabular}[c]{@{}l@{}}Traditional\\Machine Learning\\with Counter Vector\end{tabular}} &
  Decision Tree &
  85.63 &
  85.76 &
  85.20 &
  99.67 &
  99.66 &
  99.66 \\
                                   & Logistic Regression & 86.84          & 88.37          & 85.20          & 99.63          & 99.63          & 99.63          \\
                                   & Naive Bayes         & 79.04          & 63.89          & 71.47          & 98.80          & 97.69          & 97.75          \\
                                   & Random Forest       & 86.35          & 87.74          & 87.05          & \textbf{99.68} & \textbf{99.68} & \textbf{99.68} \\
                                   & SVM                 & \textbf{87.39} & \textbf{89.72} & \textbf{88.55} & 99.59          & 99.59          & 99.59          \\ \hline
\multirow{3}{*}{Transfer Learning} & mBERT               & 91.62          & 91.62          & 91.62          & 99.84          & 99.84          & 99.84          \\
                                   & DistilmBERT         & \textbf{91.63} & \textbf{91.67} & \textbf{91.65} & \textbf{99.85} & \textbf{99.85} & \textbf{99.85} \\
                                   & XLM-R               & 90.29          & 90.32          & 90.31          & 99.82          & 99.82          & 99.82          \\ \hline
\end{tabular}%
}
\end{table}

In Japanese gender detection based on names, the experimental results reveal noteworthy trends and differentiating performances among various models. In terms of the traditional machine learning approach, when applying TF-IDF data representation, Random Forest emerges as the most robust method, achieving the highest F1 score in both converted and original romaji datasets, 87.81\% and 99.66\% on average, respectively. In contrast, when Count Vectorizer is employed, Random Forest remains a formidable performer in the original romaji data with 87.05\% of the F1 score, while SVM surpasses all others in the converted data, achieving an F1 score of 88.55\%. In the realm of transfer learning, DistilmBERT gained the highest F1 score across both types of converted and original romaji input names with 91.65\% and 99.85\%, respectively, followed closely by mBERT and XLM-R with all above 90\% for all. Observations that all models trained by original romaji names slightly outperform the ones with converted names show that the accuracy of the tool we used for converting kanji into romaji still needs to be improved, but usable. 

\begin{table}[H]
\centering
\caption{The experimental results of best-performance models of each approach on first name, last name, and full name input data.}
\label{tab:separate_experimental_results}
\resizebox{\textwidth}{!}{%
\begin{tabular}{llcccccc}
\hline
\multicolumn{1}{c}{\multirow{2}{*}{\textbf{Predicted Name}}} &
  \multicolumn{1}{c}{\multirow{2}{*}{\textbf{Model}}} &
  \multicolumn{3}{c}{\textbf{Converterd}} &
  \multicolumn{3}{c}{\textbf{Original}} \\ \cline{3-8} 
\multicolumn{1}{c}{} &
  \multicolumn{1}{c}{} &
  \textbf{Female} &
  \textbf{Male} &
  \textbf{Average} &
  \textbf{Female} &
  \textbf{Male} &
  \textbf{Average} \\ \hline
\multirow{3}{*}{First Name} & RF + TF-IDF          & 88.18          & 90.25          & 89.21          & \textbf{99.81} & \textbf{99.81} & \textbf{99.81} \\
                            & SVM + Count Vector & 87.30          & 89.78          & 88.54          & 99.80          & 99.80          & 99.80          \\
                            & DistilmBERT          & \textbf{91.83} & \textbf{92.80} & \textbf{92.33} & 99.80          & 99.80          & 99.80          \\ \hline
\multirow{3}{*}{Last Name}  & RF + TF-IDF          & 15.34          & 63.83          & 39.58          & 0.80           & 66.41          & 33.61          \\
                            & SVM + Count Vector & 2.19           & \textbf{66.29} & 34.24          & 0.80           & \textbf{66.42} & 33.61          \\
                            & DistilmBERT          & \textbf{39.07} & 57.41          & \textbf{48.32} & \textbf{37.55} & 57.97          & \textbf{47.86} \\ \hline
\multirow{3}{*}{Full Name}  & RF + TF-IDF          & 87.11          & 88.50          & 87.81          & 99.67          & 99.66          & 99.66          \\
                            & SVM + Count Vector & 87.39          & 89.72          & 88.55          & 99.68          & 99.68          & 99.68          \\
                            & DistilmBERT          & \textbf{91.63} & \textbf{91.67} & \textbf{91.65} & \textbf{99.85} & \textbf{99.85} & \textbf{99.85} \\ \hline
\end{tabular}%
}
\end{table}

We next conduct experiments on input as first name and last name, compared to the above full-name experiments. Note that we only choose the best-performance model for each approach from Table \ref{tab:experimental_results} for evaluation. The obtained results in Table \ref{tab:separate_experimental_results} show that DistilmBERT consistently outperforms the other models across all data types, achieving the highest F1 score in predicting gender from first, last, and full names. However, predicting gender from last names proves more challenging for all models. The fact that the last names of Japanese males and females can be similar leads to the significantly lower F1 score achieved in gender detection based on the last name. Additionally, the performance of these models on only first-name data is comparable to full-name ones, proving that first name is the critical factor in detecting the gender of Japanese people.



\section{Conclusion and Future Work}

This paper introduced a novel dataset serving the task of gender detection based on Japanese names consisting of more than 60K rows. Moreover, we proposed Gendec, a framework comprised of various approaches, including traditional machine learning and transfer learning, to detect the biological gender of Japanese by their names. The achieved experimental results showed that the transfer learning approach, particularly with DistilmBERT, outperformed almost all other models on the task. Furthermore, the experiments proved that the differences between genders lead to the differences in Japanese first names.

In future, we plan to exploit the personal name dataset on different aspects, such as cultural inheritance characteristics, to explore the meaning of Japanese people's names. Additionally, experiments about the relation between romaji and kanji forms of Japanese names will be considerably conducted to figure out the pattern of that. 

\section*{Acknowledgement}
This research is funded by the University of Information Technology - Vietnam National University Ho Chi Minh City under grant number D1-2023-60.

\bibliographystyle{plain}
\bibliography{references} 

\end{document}